\documentclass[final]{cvpr}

\usepackage{times}
\usepackage{epsfig}
\usepackage{graphicx}
\usepackage{amsmath}
\usepackage{amssymb}
\usepackage{multirow}
\usepackage{tabularx}
\usepackage{booktabs}
\usepackage{colortbl}
\usepackage{arydshln}
\usepackage{subcaption}
\usepackage[dvipsnames]{xcolor}
\usepackage{tablefootnote}
\usepackage{enumitem}
\setlist[itemize]{leftmargin=*}

\usepackage{amsmath,amsfonts,bm}

\def\eqref#1{equation~\ref{#1}}

\def\1{\bm{1}}

\def\vzero{{\bm{0}}}
\def\vone{{\bm{1}}}

\def\vtheta{{\bm{\theta}}}
\def\vdelta{{\bm{\delta}}}
\def\vphi{{\bm{\phi}}}

\def\vv{{\bm{v}}}
\def\vw{{\bm{w}}}

\def\vy{{\bm{y}}}
\def\vz{{\bm{z}}}

\newcommand{\alphav}{{\boldsymbol \alpha}}

\DeclareMathAlphabet{\mathsfit}{\encodingdefault}{\sfdefault}{m}{sl}
\SetMathAlphabet{\mathsfit}{bold}{\encodingdefault}{\sfdefault}{bx}{n}

\def\gD{{\mathcal{D}}}

\def\gL{{\mathcal{L}}}

\def\gN{{\mathcal{N}}}

\def\gR{{\mathcal{R}}}

\newcommand{\E}{\mathbb{E}}

\newcommand{\R}{\mathbb{R}}

\usepackage{xcolor}
\definecolor{asparagus}{rgb}{0.53, 0.66, 0.42}

\newcommand{\specialcell}[2][c]{%
  \begin{tabular}[#1]{@{}c@{}}#2\end{tabular}}
\newcommand{\specialcelll}[2][l]{%
  \begin{tabular}[#1]{@{}l@{}}#2\end{tabular}}

\usepackage{tabu}

\usepackage[pagebackref=true,breaklinks=true,colorlinks,bookmarks=false]{hyperref}

\usepackage[font=small,labelfont=bf]{caption}  
\usepackage{makecell}
\usepackage{tabulary}
\definecolor{demphcolor}{RGB}{144,144,144}

\definecolor{mygray}{gray}{0.4}
\usepackage{pifont}
\newlength\savewidth

\makeatletter\renewcommand\paragraph{\@startsection{paragraph}{4}{\z@}
  {.5em \@plus1ex \@minus.2ex}{-.5em}{\normalfont\normalsize\bfseries}}\makeatother

\newcolumntype{C}[1]{>{\centering\arraybackslash}p{#1}}
\newcolumntype{R}[1]{>{\raggedleft\arraybackslash}p{#1}}
\newcolumntype{L}[1]{>{\raggedright\arraybackslash}p{#1}}

\def\cvprPaperID{10044} 
\def\confYear{CVPR 2021}

\begin{document}

\title{A Closer Look at the Robustness of Vision-and-Language Pre-trained Models}

\author{Linjie Li, Zhe Gan, Jingjing Liu  \\
Microsoft \\
{\tt\small \{lindsey.li, zhe.gan, jingjl\}@microsoft.com}
}
\maketitle

\begin{abstract}
Large-scale pre-trained multimodal transformers, such as ViLBERT and UNITER, have propelled the state of the art in vision-and-language (V+L) research to a new level. Although achieving impressive performance on standard tasks, to date, it still remains unclear how robust these pre-trained models are. To investigate, we conduct a host of thorough evaluations on existing pre-trained models over 4 different types of V+L specific model robustness: ($i$) Linguistic Variation; ($ii$) Logical Reasoning; ($iii$) Visual Content Manipulation; and ($iv$) Answer Distribution Shift. Interestingly, by standard model finetuning, pre-trained V+L models already exhibit better robustness than many task-specific state-of-the-art methods. To further enhance model robustness, we propose \textsc{Mango}, a generic and efficient approach that learns a \textbf{M}ultimodal \textbf{A}dversarial \textbf{N}oise \textbf{G}enerat\textbf{O}r in the embedding space to fool pre-trained V+L models. Differing from previous studies focused on one specific type of robustness, \textsc{Mango} is task-agnostic, and enables universal performance lift for pre-trained models over diverse tasks designed to evaluate broad aspects of robustness. Comprehensive experiments demonstrate that \textsc{Mango} achieves new state of the art on 7 out of 9 robustness benchmarks, surpassing existing methods by a significant margin. As the first comprehensive study on V+L robustness, this work puts robustness of pre-trained models into sharper focus, pointing new directions for future study.
\end{abstract}

\vspace{-5mm}
\section{Introduction}
Large-scale multimodal pre-training has taken innovative strides in the realm of vision-and-language (V+L) research~\cite{cao2020behind,lin2020interbert,lu201912,shi2020contrastive,sun2019videobert,sun2019learning,li2020hero,zhu2020actbert,li2020weakly}.
Pre-trained models~\cite{li2020oscar,yu2020ernie,su2019vl,li2019unicoder,li2019visualbert,huang2020pixel} such as  ViLBERT~\cite{lu2019vilbert}, LXMERT~\cite{tan2019lxmert} and UNITER~\cite{chen2019uniter}
have demonstrated great generalizability over diverse V+L tasks~\cite{zhou2019unified,hu2020vivo,murahari2019large,wang2020vd,alberti2019fusion,hao2020towards,majumdar2020improving,cao2020behind}, such as
Visual Question Answering (VQA)~\cite{antol2015vqa}, 
Visual Commonsense Reasoning (VCR)~\cite{zellers2019recognition}, and  Referring Expression Comprehension~\cite{yu2016modeling}.
However, these benchmarks usually possess similar data distribution between training and test sets, with little-to-none linguistic variation in textual queries, and use only clean natural images without any visual content manipulation. Therefore, although effective for general model evaluation, these standard benchmarks lack the ability to explicitly evaluate model \emph{robustness}\footnote{We do not focus on \emph{adversarial} robustness in this work, as no existing adversarial benchmark is available. 
Instead, we investigate existing robustness benchmarks, which are designed with challenging settings and validated by human judges. See Section~\ref{sec:benchmark} for a more detailed discussion.}. 

 To conduct a full dissection on model robustness, we launch a thorough investigation with systematic evaluations of V+L models over 4 generic types of robustness: ($i$) robustness against \emph{linguistic variation}; 
($ii$) robustness against \emph{logical reasoning}; ($iii$) robustness against \emph{visual content manipulation}; and
($iv$) robustness against \emph{answer distribution shift}
between training and test splits.
Given the abundance of diverse datasets and splits on the popular VQA task, we take VQA as the focal point of our investigation, and compile 
an assemblage of 9 diverse VQA datasets that cover each type of model robustness: ($i$) VQA-Rephrasings~\cite{shah2019vqa-rephrase} for \emph{linguistic variation}; 
($ii$) VQA-LOL (Compose and Supplement)~\cite{gokhale2020vqa-lol}, VQA-Introspect~\cite{selvaraju2020vqa-introspect} and GQA~\cite{hudson2019gqa} for \emph{logical reasoning};
($iii$) IV-VQA and CV-VQA~\cite{agarwal2020causal-vqa} for \emph{visual content manipulation}; and ($iv$) VQA-CP v2~\cite{agrawal2018vqa-cp} and GQA-OOD~\cite{kervadec2020gqa-ood} for \emph{answer distribution shift}.  To the best of our knowledge, this is the first systematic evaluation of pre-trained V+L models through the lens of \emph{robustness}. Interestingly, analysis on two archetypal V+L models reveals that by standard finetuning, pre-trained models already exhibit better robustness than many task-specific state-of-the-art methods. However, the achieved robustness is still limited, and far from human performance.

\begin{figure*}
     \centering
     \begin{subfigure}[b]{0.62\textwidth}
         \centering
         \includegraphics[width=\textwidth]{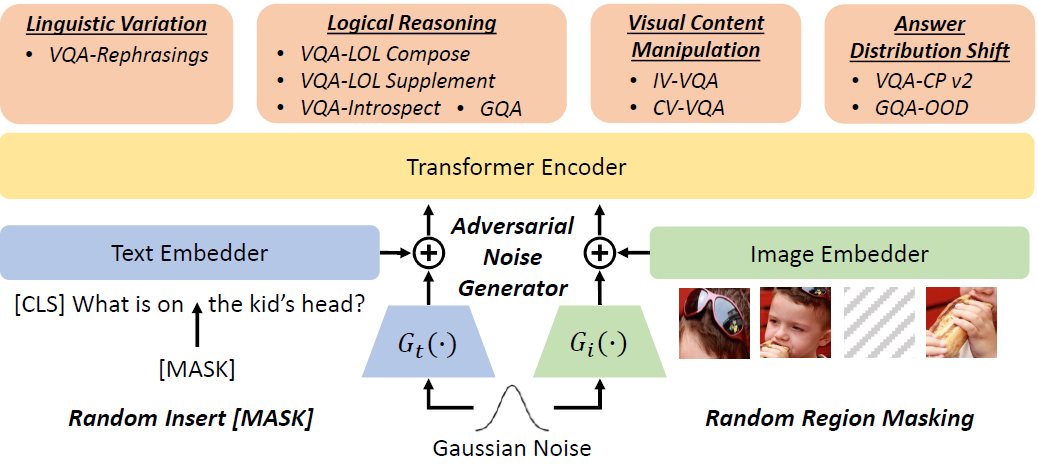}
         \caption{\small{Overview of \textsc{Mango}, which trains adversarial noise generators to add perturbations at embedding level. Random masking on image and text inputs are designed to promote more diverse adversarial embeddings.}}
         \label{fig:model}
     \end{subfigure}
     \hfill
     \begin{subfigure}[b]{0.33\textwidth}
         \centering
         \includegraphics[width=\textwidth]{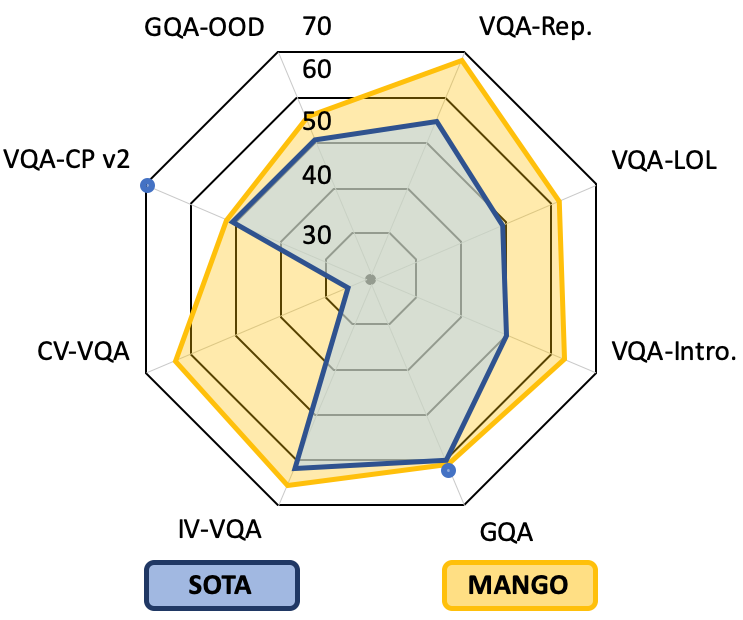}
         \caption{\small{Comparison between \textsc{Mango} and SOTA. The \textcolor{blue}{blue dots} represent methods exploiting additional task-specific information.\footnotemark}}
         \label{fig:poly}
     \end{subfigure}
    \vspace{-8pt}
    \caption{\small {Illustration of the proposed \textsc{Mango} framework and performance comparison between \textsc{Mango} and SOTA. }}
    \label{fig:main}
    \vspace{-3mm}
\end{figure*}

Recently, adversarial training (AT)~\cite{tramer2017ensemble,xie2019feature,shafahi2019adversarial,pang2020boosting,zhang2019theoretically,wong2020fast} has shown success
on \emph{standard} V+L tasks~\cite{gan2020large,tang2020semantic}. Inspired by this, we investigate whether AT can also serve as an effective conduit to improve performance on robustness benchmarks aforementioned. Our evaluation of \textsc{Villa}~\cite{gan2020large} (AT-enhanced pre-trained model) 
shows that by injecting adversarial perturbation to multimodal embeddings, PGD-based (Projected Gradient Descent) AT~\cite{madry2017towards,  xie2020adversarial,zhu2019freelb,jiang2019smart,liu2020adversarial,kong2020flag} can help the model adapt to linguistic variation and visual content manipulation, yielding better model robustness; but with only limited effect (sometimes even hurting model performance) 
on datasets that exhibit salient data distribution gap between training and test sets (\emph{e.g.}, VQA-CP v2, GQA-OOD).

\footnotetext{LMH~\cite{tramer2017ensemble} and MMN~\cite{chen2021meta} on VQA-CP v2 and GQA are used to plot the SOTA polygon for fair comparison. For CV-VQA and IV-VQA, performance is computed as $100 - $\#flips and $100 - 5\times$\#flips, respectively. VQA-LOL performance is the average of accuracies on VQA-LOL Compose and VQA-LOL Supplement.}

To achieve better robustness across all aspects, we propose \textsc{Mango}~\includegraphics[height=10pt]{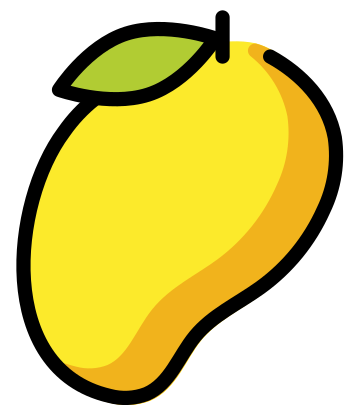} (\textbf{M}ultimodal \textbf{A}dversarial \textbf{N}oise \textbf{G}enerat\textbf{O}r),
a generic and efficient approach that introduces adversarial noise to multimodal embedding space for robustness enhancement. As shown in Figure~\ref{fig:model}, instead of relying on PGD to generate adversarial perturbation, \textsc{Mango} learns an adversarial noise generator in the form of a trained neural network to fool the model. Following~\cite{gan2020large}, perturbation is added to the embedding space for all modalities, as our goal is the \emph{end results} of AT, rather than crafting actual adversarial examples.
Compared with PGD-based training which is time-consuming, \textsc{Mango} is lightweight and efficient, without the repetitive iterations of gradient calculation required in PGD-based approach.

To enable diverse adversarial embeddings, we further propose to randomly mask image regions and randomly insert \texttt{[MASK]} tokens when adding adversarial noise to image and word embeddings. 
Empirical results show that \textsc{Mango} significantly improves model robustness across all tasks considered, 
compared to PGD-based methods.

Our main contributions are summarized as follows. ($i$) We conduct the first known study to systematically examine model robustness of prevailing V+L pre-trained models.
($ii$) We propose \textsc{Mango}, a generic and efficient adversarial training approach to enhance V+L model robustness. 
($iii$) As summarized in Figure~\ref{fig:poly}, we achieve new state of the art on 7 out of 9 robustness benchmarks, lifting model performance by \emph{+11.74} on VQA-Rephrasings, \emph{+12.50} on VQA-LOL Compose, \emph{+9.96} on VQA-LOL Supplement, \emph{+12.55} on VQA-Introspect,  \emph{+0.84} on IV-VQA, \emph{+42.92} on CV-VQA, and \emph{+3.70} on GQA-OOD.

\begin{table*}[t!]
    \centering
\small
\resizebox{\textwidth}{!}{
    \small
    \begin{tabu}{llccccccccc}
    \hline
    \multirow{2}{*}{\specialcelll{Type}} & \multirow{2}{*}{Benchmark}  & \multirow{2}{*}{Metric} & \multirow{2}{*}{Q Type}  & \multicolumn{3}{c}{Train}  & \multicolumn{2}{c}{Val} & \multicolumn{2}{c}{Test}\\
    \cmidrule(lr){5-7}  \cmidrule(lr){8-9} \cmidrule(lr){10-11}

    & & & & Source & \#IQ & len(Q) & \#IQ & len(Q)& \#IQ & len(Q)\\
    \hline
    Lingual & VQA-Rep.~\cite{shah2019vqa-rephrase} & Acc. & All &  VQA v2~\cite{goyal2017making} train &  444K  & 6.20  & 162K & 7.15 & - & -\\
    \hline
    \multirow{4}{*}{ Reason } & VQA-LOL Comp.~\cite{gokhale2020vqa-lol} & Acc. & Y/N &  VQA v2 train &   444K  & 6.20 & 43K & 12.09 & 291K & 12.12  \\
    & VQA-LOL Supp.~\cite{gokhale2020vqa-lol} & Acc. & Y/N &  VQA v2 train &   444K  & 6.20 & 9K & 15.15 & 669K & 15.19 \\
    & VQA-Intro.~\cite{selvaraju2020vqa-introspect} & \footnotesize{M$\checkmark$S$\checkmark$} & All & VQA v1~\cite{antol2015vqa} train & 248K & 6.21 & - & - & 95K & 6.36  \\
    & GQA~\cite{hudson2019gqa} & Acc. & All & - & 943K & 8.76 & 132K & 8.77 & 13K & 8.51 \\
    \hline
    \multirow{2}{*}{ \specialcelll{Visual} } & IV-VQA~\cite{agarwal2020causal-vqa} & \#flips & All & VQA v2 train &   444K  & 6.20  &   120K & 5.85 & - & - \\
    & CV-VQA~\cite{agarwal2020causal-vqa}  & \#flips & Num. &  VQA v2 train &   444K  & 6.20 & 4K &  5.83 & - & - \\
    \hline
    \multirow{2}{*}{ \specialcelll{Answer} } & VQA-CP v2~\cite{agrawal2018vqa-cp} & Acc.  & All & - & 438K & 6.14  & - &- & 220K & 6.31\\
    & GQA-OOD~\cite{kervadec2020gqa-ood} & Acc. & All & GQA train & 943K & 8.76 &  51K & 8.09 & 3K & 7.70\\
    \hline
    \end{tabu}
}
\vspace{-8pt}
    \caption{\small{Detailed descriptions of each downstream benchmark, including robustness type, evaluation metric, question type, training data source and statistics on train, val, test data in terms of number of Image-Question pairs (\#IQ) and average question length (len(Q)). We use the training data provided with the benchmark unless specified otherwise. Results on val split are reported when test split is not available. Acc. is short for Accuracy.  M$\checkmark$S$\checkmark$ is a consistency measure between main questions and sub-questions in VQA-Introspect. \#flips is the number of predictions mismatched before and after visual content manipulation.}}
    \label{tab:dwonstream_data}
    \vspace{-3mm}
\end{table*}

\section{Robust VQA} \label{sec:benchmark}
\paragraph{Terminology} We start with definition of the terminology we use throughout the paper. We follow VQA literature~\cite{cadene2019rubi,wu2019self,teney2020value,gokhale2020mutant,kervadec2020gqa-ood} to unify different forms of challenging bias and out-of-distribution generalization as \emph{robustness}, which is different from its definition in adversarial machine learning. Robustness does not always mean ``adversarial robustness'' in literature, \emph{e.g.}, it can also refer to model robustness towards common image corruptions~\cite{rusak2020simple,zhang2019making,hendrycks2019benchmarking}. In the language of adversarial machine learning, our definition of robustness here can be unserstood as the ``generalization" performance on the challenging robust VQA benchmarks. 

\vspace{1.5mm}
\noindent\textbf{Existing Benchmarks}\, There has been a few independent studies on V+L robustness, mostly focusing on variations of the popular VQA task. 
VQA-CP \cite{agrawal2018vqa-cp}, drawn from VQA v2 dataset~\cite{goyal2017making}, is the first benchmark proposed to evaluate (and reduce) question-oriented language bias in VQA models. Considerable effort~\cite{agrawal2018vqa-cp,jing2020overcoming,kv2020reducing,niu2020counterfactual,selvaraju2019taking, abbasnejad2020counterfactual} has been invested on VQA-CP along 3 dimensions: ($i$)  compensating for question-answer distribution patterns through a regularizer based on an auxiliary model~\cite{niu2020counterfactual, cadene2019rubi,clark2019don,teney2020unshuffling, grand2019adversarial, jing2020overcoming}; ($ii$) taking advantage of additional supervision from human-generated attention maps~\cite{selvaraju2019taking,wu2019self,gokhale2020mutant}; and ($iii$) synthesizing counterfactual examples to augment training set~\cite{abbasnejad2020counterfactual, chen2020counterfactual,teney2020learning}. 
Recent work~\cite{teney2020value} shows that simple methods such as generating answers at random can already surpass state of the art on some question types. The recent GQA-OOD~\cite{kervadec2020gqa-ood}, another robustness-focused task, is designed based on a fine-grained reorganization of the original GQA dataset~\cite{hudson2019gqa}.

Besides answer distribution shift, other types of VQA model robustness are also studied: VQA-Rephrasings~\cite{shah2019vqa-rephrase} exposes the brittleness of VQA models to linguistic variations in questions, and proposes cyclic consistency to improve robustness; \cite{ray2019sunny} tackles antonym consistency; \cite{agarwal2020causal-vqa} studies robustness against automated semantic image manipulations, and tests for prediction consistency to questions on clean images and corresponding manipulated images.

Further studies investigate robustness against logical reasoning. For instance, \cite{selvaraju2020vqa-introspect}  provides a dataset containing perception-related sub-questions per question for a new reasoning split of VQA dataset.
\cite{gokhale2020vqa-lol} constructs VQA-LOL with questions containing logical compositions and linguistic transformations to examine model ability in logical reasoning. GQA~\cite{hudson2019gqa} also falls into this category, as its large-scale rule-based questions support analysis on different reasoning skills of VQA models.

Despite the continuous effort in enhancing robustness of VQA models, these works mostly focus on either task-specific models or a single type of robustness. To provide the first comprehensive study on V+L robustness, we compile a full list of existing datasets, and group them into four robustness types: \emph{Lingual},  \emph{Visual}, \emph{Reason}, and \emph{Answer} (Table~\ref{tab:dwonstream_data}). Covering various respects of a `stress test' for V+L models, from \emph{linguistic} to \emph{visual} variations, from \emph{reasoning complexity} to \emph{answer distribution}, this compilation serves as a unified yardstick for evaluating V+L model robustness and a guidance for future study on robust model design\footnote{Although the benchmarks considered here are VQA tasks, these robustness types are generic, and can be extended to other V+L tasks as well. For example, \emph{Lingual} and \emph{Visual} are naturally applicable to any task with text and image inputs. \emph{Reason} and \emph{Answer} benchmarks can be constructed via logical combination of text inputs or reshuffling train/val/test splits.}. As a start, we introduce a generic and effective approach that can lift model performance over all types of V+L robustness indiscriminately, which will be discussed in next section. 

\section{\textsc{Mango} Framework}
In Sec.~\ref{sec:background}, we briefly review V+L pre-training. Sec.~\ref{sec:baseline} introduces a simple baseline that injects Gaussian noise.
The proposed \textsc{Mango} approach is detailed in Sec.~\ref{sec:mango}.

\subsection{Multimodal Pre-training} \label{sec:background}
During pre-training, a model takes image-text pairs as input, where the image input is usually fed into a fixed object detector~\cite{anderson2018bottom} to extract a set of region-level features $\vv = \{\vv_1, ..., \vv_K\}$ ($\vv_i \in \R^{d_v}$).
The text input is tokenized and projected into high-dimensional feature vectors $\vw = \{\vw_1, ..., \vw_L\}$ through a learnable word embedding layer ($\vw_i \in \R^{d_w}$). 
These embeddings, along with the location/position embeddings of regions and tokens, are then fed into a multi-layer transformer to obtain contextualized representations for each image region and word token, through well-designed pre-training tasks such as masked language modeling and image-text matching. 
Formally, the above process can be abbreviated as $\vz_{v}, \vz_{w}, \vz_{cls}= f_{\vtheta}(\vv, \vw)$, where $\vtheta$ denotes all the trainable parameters. $\vz_{v}\in \R^{K\times d_v}$ and $\vz_{w}\in \R^{L\times d_w}$ represent the final representations for image regions and word tokens, respectively. $\vz_{cls}$, the embedding of the special token \texttt{[CLS]}, which is often inserted to the beginning of image-text input, is considered as the joint image-text representation used for downstream finetuning. 
A detailed literature review on V+L pre-training is provided in Appendix.

Take VQA task as an example. Given an image-question pair $(\vv, \vw)$ in dataset $\gD$, the goal is to predict an answer that best matches ground-truth answer $\vy$. When finetuning on VQA task, a multi-layer perceptron (MLP) layer is added on top of $\vz_{cls}$, and Binary Cross Entropy (BCE) loss is used to supervise model training. The finetuning process can be formulated as an empirical risk minimization problem:
\begin{equation}
     \min_{\vtheta}  \underset{(\vv, \vw,\vy)\sim \gD}{\E} [ \gL_{\text{BCE}}(f_{\vtheta}(\vv, \vw),\vy)]\,.
\end{equation}

\subsection{Gaussian Noise Augmentation} \label{sec:baseline}
Randomized smoothing~\cite{duchi2012randomized} advocates the addition of random perturbations to model inputs, which can often yield better model performance. 
Recent study~\cite{rusak2020simple} also shows that perturbing clean images with Gaussian noise is effective in improving model robustness against image corruptions for image classification. Inspired by this, we use Gaussian noise augmentation as a simple baseline to investigate model robustness under V+L setting. Instead of adding noise to raw image pixels as in~\cite{rusak2020simple}, we add perturbations directly to the embeddings: 
\begin{equation}
     \min_{\vtheta} \underset{(\vv, \vw,\vy)\sim \gD}{\E} \, \underset{\vdelta_v\sim \gN(\vzero,\sigma^2\vone)}{\E}[ \gL_{\text{BCE}}(f_{\vtheta}(\vv+\vdelta_v, \vw),\vy)]\,,
\end{equation}
where $\sigma$ is the standard deviation of Gaussian noise.  Similarly, we add Gaussian noise to the word embeddings: 
\begin{equation}
     \min_{\vtheta} \underset{(\vv, \vw,\vy)\sim \gD}{\E} \, \underset{\vdelta_w\sim \gN(\vzero,\sigma^2\vone)}{\E}[ \gL_{\text{BCE}}(f_{\vtheta}(\vv, \vw+\vdelta_w),\vy)]\,. \nonumber
\end{equation}

\subsection{Adversarial Noise Generator} \label{sec:mango}
Adding Gaussian noise to clean image-text pairs can augment training examples to a certain level. However, as the training continues, the model can gradually adapt to the perturbations which are sampled from the \emph{same} Gaussian noise distribution. To produce \emph{harder} perturbations that can fool the backbone network such as UNITER~\cite{chen2019uniter} or LXMERT~\cite{tan2019lxmert}, we propose to actively learn an adversarial noise generator.
Specifically, we aim to discover an adversarial noise distribution, from which the sampled noises, when added to the multimodal embeddings, can maximally confuse the backbone network. Note that our goal is not to model the explicit density form of such a distribution, as we only care about the noise samples drawn from the distribution. To achieve this, the adversarial noise generator takes in Gaussian noise samples as input, and produces adversarial noise samples through a learned neural network. 

Take image modality as an example. Let $g_{\vphi_v}: \R^{d_v} \rightarrow \R^{d_v}$ denote the adversarial noise generator.  The adversarial noise $\vdelta_v$ is generated by $\vdelta_v = g_{\vphi_v}(\alphav), \alphav \in \gN(\vzero, \vone)$. 
Intuitively, to maximally fool the backbone network, we want to maximize prediction errors on these adversarially perturbed samples. In the meantime, we want the model to possess less confidence in its predictions on perturbed samples than clean samples, to promote harder adversarial examples   
Therefore, the objective of the adversarial noise generator is to \emph{maximize} the sum of two losses: ($i$) task-specific loss (\eg, BCE loss for VQA task); ($ii$) adversarial loss (\eg, BCE loss for adversarial data, and the Kullback-Leibler (KL) divergence loss between the predicted answer distribution of perturbed samples and that of clean samples). On the other hand, the trained model aims to minimize both losses by taking adversarial embeddings as data augmentation. Formally, the min-max game can be defined as: 
\begin{align}
    \min_{\vtheta} \max_{\vphi_v} &\underset{(\vv, \vw,\vy)\sim \gD}{\E} \, \underset{\alphav \in \gN(\vzero, \vone)}{\E}[ \gL_{std}(\vtheta, \vphi_v) + \beta\gR_{at}(\vtheta, \vphi_v)]\, , \nonumber
\end{align}
where $\beta$ is a hyper-parameter, and
\begin{align}
    \gL_{std}(\vtheta, \vphi_v) &= \gL_{\text{BCE}}(f_{\vtheta}(\vv, \vw),\vy)\,, \\ \nonumber
    \gR_{at}(\vtheta, \vphi_v) & = \gL_{\text{BCE}}(f_{\vtheta}(\vv+g_{\vphi_v}(\alphav), \vw),\vy) \\
    & +\gL_{kl}(f_{\vtheta}(\vv+g_{\vphi_v}(\alphav), \vw), f_{\vtheta}(\vv, \vw)),
\end{align}
where $\gR_{kl}(p,q) = \text{KL}(p||q)+\text{KL}(q||p)$, $p,q$ denote two probability distributions. The first term in $\gR_{at}(\vtheta, \vphi_v)$ promotes label-preserving adversarial perturbations; while the second term advocates more fine-grained label preservation, meaning that the probability distribution across all answers is used as \emph{soft} label, instead of using the ground truth answer index as hard label. Similarly, we learn an adversarial noise generator (with parameters $g_{\vphi_w}$) that corresponds to the text modality, jointly trained with $g_{\vphi_v}$.\footnote{The corresponding equations are omitted for simplicity.}

During training, we alternate between an outer loop of the backbone network update and an inner loop of generator update. We constrain the noise samples $\vdelta_v$ and $\vdelta_w$ to be within the sphere $||\vdelta_v||_2 = ||\vdelta_w||_2 = \epsilon$, by scaling the generator output with a scalar. $\epsilon$ is set as 1 in all our experiments, as perturbations with a smaller norm is more likely to preserve original semantics. 
For better efficiency, we also accumulate the gradients of adversarial noise generator, and only update the generator's parameters every $T$ times (set to 20 or 40 in experiments) of backbone update.

The proposed adversarial noise generator is lightweight, consisting of only a few linear layers.  Such a light model can easily trap in local minimum when competing with a deep backbone network. Thus, at regular intervals, we replace the learned noise generator with a new one trained from scratch. Each time, the new generator is trained against the latest learned parameters of the backbone.

\vspace{5pt}
\noindent\textbf{Random Masking}\, 
Adversarial noise generator, although produces more challenging and more diverse noise perturbations, does not alter the intrinsic statistics of training examples, such as the distribution of question lengths and image regions. In practice, we observe significant mismatch in these statistics between training and test splits of robustness benchmarks. For example, the average length of questions in VQA-LOL~\cite{gokhale2020vqa-lol} test split is 2-3 times longer than that in VQA v2~\cite{goyal2017making} training split. The region distribution of images in IV-VQA and CV-VQA~\cite{agarwal2020causal-vqa} is very different from VQA v2 training split, due to visual content manipulation. To compensate for such statistic mismatch, we propose to randomly mask image regions (by zeroing out corresponding feature vectors) as well as randomly insert \texttt{[MASK]} tokens when adding adversarial noise to image and word embeddings. Empirically, this simple technique is effective in further boosting model robustness.

\begin{table*}[t!]
    \centering
\resizebox{\textwidth}{!}{
    \begin{tabu}{>{\small}p{0.01\textwidth} lc  cccccccccc}
\hline
\rowfont{\small}%
& &  & Lingual & \multicolumn{4}{c}{Reason} & \multicolumn{2}{c}{Visual} & \multicolumn{2}{c}{Answer} &  \\
 \cmidrule(lr){4-4} \cmidrule(lr){5-8}   \cmidrule(lr){9-10} \cmidrule(lr){11-12}
\rowfont{\small}%
& \normalsize{Model}  & & \specialcell{VQA-\\Rep.} & \specialcell{VQA-LOL\\Comp.} & \specialcell{VQA-LOL\\Supp.} & \specialcell{VQA-\\Intro.} & \specialcell{GQA}  & \specialcell{IV-\\VQA}  & \specialcell{CV-\\VQA}  & \specialcell{VQA-\\CP v2} & \specialcell{GQA-\\OOD} & \specialcell{VQA\\v2} \\

 \cmidrule(lr){4-4} \cmidrule(lr){5-5} \cmidrule(lr){6-6}   \cmidrule(lr){7-7} \cmidrule(lr){8-8}   \cmidrule(lr){9-9} \cmidrule(lr){10-10} \cmidrule(lr){11-11} \cmidrule(lr){12-12} \cmidrule(lr){13-13} 

\rowfont{\small}%
& & Meta-Ave. \textcolor{green}{$\uparrow$}
& Acc. \textcolor{green}{$\uparrow$} 
& Acc. \textcolor{green}{$\uparrow$} 
& Acc. \textcolor{green}{$\uparrow$} 
& M$\checkmark$ S$\checkmark$ \textcolor{green}{$\uparrow$} 
& Acc. \textcolor{green}{$\uparrow$} 
& \#flips \textcolor{red}{$\downarrow$} 
& \#flips \textcolor{red}{$\downarrow$} 
& Acc. \textcolor{green}{$\uparrow$} 
& Acc. \textcolor{green}{$\uparrow$} 
& Acc. \textcolor{green}{$\uparrow$}\\ 

\hline
1 & SOTA & N/A & 56.59 & 48.99 & 50.54 & 50.05  & \textbf{63.17} & 7.53 & 78.44 & \textbf{69.52} & 52.70 & \textbf{74.69} \\
\hline
2 & \textsc{Uniter}$_\text{B}$ & 40.98 & 64.56 & 54.54 & 50.00 & 56.80  & 59.99  & 8.47 & 40.67 & 46.93 & 53.43 & 72.70\\
3 & \textsc{Mango}$_\text{B}$ & 42.80 & 65.80 & 56.22 & 56.49 & 58.33 & 60.65 & 7.32 & 38.11 & 47.52  & 55.15 & 73.24\\
\hline
4 & \specialcelll{\textsc{Villa}$_\text{B}$}&42.37 & 65.35 & 54.90 & 56.17 & 58.29  & 60.26  & 7.07 & 38.28 & 46.39 & 54.11 & 73.59 \\
5 & \specialcelll{\textsc{Mango}$_\text{VB}$}& 43.08 & 65.91 & 55.44 & 57.58 & 58.94  & 60.73  & 7.43 & 38.25& 48.63 & 55.79 & 73.45\\
\hline
6 & \textsc{Uniter}$_\text{L}$  & 43.37 & 67.64 & 58.60 & 55.95 & 57.64  & 60.30  & 8.20 & 36.66 & 50.98 & 53.65 & 73.82\\
7 & \textsc{Mango}$_\text{L}$  & \underline{45.27} &\textbf{68.33} & \underline{59.45} & \textbf{60.50}& \underline{62.14}  & 61.10  & \textbf{6.69} & \textbf{35.52} & \underline{52.76} & \textbf{56.40} & \underline{74.26}\\
\hline
8 & \specialcelll{\textsc{Villa}$_\text{L}$} & 44.33& 68.16 & 58.66 & 58.29 & 62.00 & 61.38  & \underline{6.70} & 37.55 & 49.10 & 55.26  & \textbf{74.69} \\
9 & \specialcelll{\textsc{Mango}$_\text{VL}$} & \textbf{45.31} & \underline{68.27} & \textbf{61.49} & \underline{58.83}& \textbf{62.60}  & \underline{61.41}  & 6.73 & \underline{35.64} & 52.55 & \underline{56.08} & 74.20\\
\hline
\end{tabu}
}
\vspace{-8pt}
    \caption{\small{Comparison to task-specific state-of-the-art (SOTA), \textsc{Uniter}, \textsc{Villa} on 9 robustness downstream benchmarks and a standard VQA benchmark.  Results are reported on val split of VQA-Rephrasings (VQA-Rep.), VQA-LOL Compose (Comp.) and Supplement (Supp.), VQA-Introspect (VQA-Intro.), IV-VQA, CV-VQA,  VQA-CP v2 and test-dev split of GQA, GQA-OOD and VQA v2. \textcolor{green}{$\uparrow$} (\textcolor{red}{$\downarrow$}) indicate the higher (lower) the better. }}
    \label{tab:sota_res}
    \vspace{-3mm}
\end{table*}

\vspace{5pt}
\noindent\textbf{Comparison with PGD-based AT}\,
Although \textsc{Mango} is similar to \textsc{Villa}~\cite{gan2020large}
in terms of learning adversarial perturbations, they are different in the sense that \textsc{Mango} learns an adversarial noise generator to generate adversarial perturbations, instead of relying on PGD as in \textsc{Villa}. 
This makes \textsc{Mango} more efficient, as computing gradients of a generic lightweight noise generator is less time-consuming. Empirically, \textsc{Mango} also achieves better performance. The comparison on model performance and training time difference is provided in Sec.~\ref{sec:main_results}. A detailed literature review on AT is provided in Appendix.

\vspace{5pt}
\noindent\textbf{Comparison with ANT}\, In ANT~\cite{rusak2020simple}, a similar noise generator is proposed to make neural networks robust
against diverse image corruptions. However, there are two key distinctions. First, we focus on transformer models for V+L tasks, whereas \cite{rusak2020simple} focuses on convolutional networks for image classification. Second, we propose to generate adversarial noise over the embeddings of images and words, while \cite{rusak2020simple} adds adversarial noise directly on image pixels.
    

\vspace{-1mm}
\section{Experiments}
\vspace{-1mm}

In experiments, we first use \textsc{Uniter}~\cite{chen2019uniter} as the backbone 
on diverse V+L tasks, then 
compare \textsc{Mango} with \textsc{Uniter} and \textsc{Villa}~\cite{gan2020large} baselines over all 9 robustness datasets (Sec.~\ref{sec:benchmark}), plus a VQA-v2 dataset. We focus our study on these 10 benchmarks as there is no existing robustness dataset on other tasks except for VQA. 


\vspace{-1mm}
\subsection{Experimental Setting}
\vspace{-1mm}

Note that almost all 10 benchmarks provide their own training split. We follow the original papers to test model robustness under the most challenging setting (shown in Table~\ref{tab:dwonstream_data}), which is to evaluate models trained on the VQA training split for VQA-Rephrasings, VQA-LOL, VQA-Introspect, IV-VQA and CV-VQA. A more detailed description of all the benchmarks are provided in Appendix.

For thorough evaluation, we compare model performance on the following set of competing methods:
\begin{itemize}
    \vspace{-1mm}
    \item \textbf{SOTA} (\emph{task-specific}): Cycle Consistency+BAN~\cite{shah2019vqa-rephrase} for VQA-Rephrasings, LOL~\cite{gokhale2020vqa-lol} for VQA-LOL Compose and Supplement, Pythia~\cite{selvaraju2020vqa-introspect,jiang2018pythia} for VQA-Introspect, NSM~\cite{hudson2019nsm} for GQA, SAAA~\cite{agarwal2020causal-vqa,kazemi2017show} for CV-VQA and IV-VQA,  MUTANT~\cite{gokhale2020mutant} for VQA-CP v2, MMN~\cite{chen2021meta,kervadec2020gqa-ood} for GQA-OOD,  \textsc{Villa}~\cite{gan2020large}
    for VQA v2; 
    \vspace{-2mm}
    \item \textbf{\textsc{Uniter}$_\text{B}$} and \textbf{\textsc{Uniter}$_\text{L}$}: standard finetuning of  \textsc{Uniter} model with base and large size, respectively;
    \vspace{-2mm}
    \item \textbf{\textsc{Villa}$_\text{B}$} and \textbf{\textsc{Villa}$_\text{L}$}: adversarial pre-trained and finetuned \textsc{Uniter} with base and large size;
    \vspace{-2mm}
    \item \textbf{\textsc{Mango}$_\text{B}$} and \textbf{\textsc{Mango}$_\text{L}$}: applying adversarial noise generator on pre-trained \textsc{Uniter}, base and large size;
    \vspace{-2mm}
    \item \textbf{\textsc{Mango}$_\text{VB}$} and \textbf{\textsc{Mango}$_\text{VL}$}: applying adversarial noise generator on adversarial pre-trained \textsc{Uniter} model (provided in the \textsc{Villa} paper~\cite{gan2020large}) with base and large size.
\end{itemize}

\subsection{Experimental Results}\label{sec:main_results} 
Table~\ref{tab:sota_res} presents the results of \textsc{Uniter}, \textsc{Villa} and \textsc{Mango} on all robustness benchmarks. Meta-Ave (average of scores across all benchmarks) is used as the global metric.\footnote{For IV-VQA and CV-VQA, we take the negative of the number of flips for calculating Meta-Ave.}
L2-5 in Table~\ref{tab:sota_res} show the performance of all the models with base size (12 layers). \textsc{Uniter}$_\text{B}$ (L2) establishes a strong baseline on different types of robustness benchmarks, with a Meta-Ave of 40.98.  \textsc{Villa}$_\text{B}$ (L4) further improves over this strong baseline by +1.39 Meta-Ave (42.37) via PGD-based adversarial training. 

\textsc{Mango}$_\text{B}$ achieves across-the-board performance lift on all robustness benchmarks over \textsc{Uniter}$_\text{B}$, harnessing an absolute gain of +1.82 on Meta-Ave. Results on VQA-v2 show that \textsc{Mango}$_\text{B}$ also boosts performance on standard VQA benchmark. 
As \textsc{Villa}$_\text{B}$ performs adversarial training on both pre-training and finetuning stages, we apply our method to their adversarial pre-trained model for fair comparison. \textsc{Mango}$_\text{VB}$ (L5) outperforms \textsc{Villa}$_\text{B}$  on 7 out of 9 robustness benchmarks, with an absolute gain +0.71 on Meta-Ave. Lastly, we compare the training speed of \text{Mango}$_\text{VB}$ and \textsc{Villa}$_\text{B}$ under the same experimental setting.\footnote{Experiments are conducted with the same batch size, gradient accumulation steps and number of GPUs.} Our experiments show that \text{Mango}$_\text{VB}$ is 25\% faster than \textsc{Villa}$_\text{B}$ (1.44 \emph{vs}. 1.92 second/step). This indicates that \textsc{Mango} is a more efficient adversarial training approach than \textsc{Villa}, thanks to its use of global noise generator instead of iterative PGD steps as used in \textsc{Villa}.

\vspace{5pt}
\noindent\textbf{Scaling Up to Large Model Size (24 Layers)}\,
Compared to base models (L2\&L4), large models (L6\&L8) have more advantage on Meta-Ave (\textsc{Uniter}: 43.37(L) vs. 40.98(B); \textsc{Villa}: 44.33(L) vs. 42.37(B)), which is consistent with the observations on standard V+L benchmarks in~\cite{chen2019uniter, gan2020large}.
When applying adversarial noise to large backbone models (L7\&L9), \textsc{Mango} further pushes the margins of performance gain across all benchmarks: an absolute gain of +1.90 over \textsc{Uniter}$_\text{L}$ and +0.98 over \textsc{Villa}$_\text{L}$ on Meta-Ave.

\vspace{5pt}
\noindent\textbf{End-to-end Comparison with SOTA}\,
\textsc{Mango} achieves new state of the art on 7 out of 9 benchmarks, in most cases surpassing existing methods by a significant margin. Specifically, \textsc{Mango} pushes state-of-the-art performance by \emph{+11.74} on VQA-Rephrasings, \emph{+12.50} on VQA-LOL Compose, \emph{+9.96} on VQA-LOL Supplement, \emph{+12.55} on VQA-Introspect,  \emph{+0.84} on IV-VQA, \emph{+42.92} on CV-VQA, and \emph{+3.70} on GQA-OOD. Best results are achieved by \textsc{Mango}$_\text{L}$ or \textsc{Mango}$_\text{VL}$.

On VQA-CP v2 and GQA, although \textsc{Mango} outperforms \textsc{Uniter} and \textsc{Villa}, there is still a gap when compared to SOTA models. SOTA methods on these two benchmarks exploit additional task-specific information. Specifically, MUTANT~\cite{gokhale2020mutant} for VQA-CP v2 is trained with excessive additional image-question pairs designed to promote positive bias; while NSM~\cite{hudson2019nsm} for GQA takes advantage of additional scene graph annotations, which are only provided in GQA. As the goal of our proposed method is to bring universal performance lift on all robustness benchmarks, we do not exploit these additional task-specific information introduced by MUTANT and GQA.

\subsection{A Closer Look into Robustness}\label{sec:robust_dicussion}

\begin{table*}[t!]
    \centering
\small
\begin{tabu}{lcccc || cccccc}
\hline
\rowfont{\small}%
\multirow{2}{*}{ Model } &  \multicolumn{4}{c||}{ VQA-Rephrasings } & VQA-Reas. & \multicolumn{5}{c}{VQA-Introspect}\\
\cmidrule(lr){2-5} \cmidrule(lr){6-6} \cmidrule(lr){7-11}
\rowfont{\footnotesize}%
&  CS(1) \textcolor{green}{$\uparrow$} & CS(2) \textcolor{green}{$\uparrow$} & CS(3) \textcolor{green}{$\uparrow$} & CS(4) \textcolor{green}{$\uparrow$} & Acc. \textcolor{green}{$\uparrow$}& M$\checkmark$ S$\checkmark$ \textcolor{green}{$\uparrow$}& M$\checkmark$ S$\times$ \textcolor{red}{$\downarrow$} & M$\times$ S$\checkmark$ \textcolor{red}{$\downarrow$} & M$\times$ S$\times$ \textcolor{red}{$\downarrow$} &
S$\checkmark|\text{M}\checkmark$ \textcolor{green}{$\uparrow$}\\
\hline
SOTA & 65.77 & 56.94 & 51.76 & 48.18 & 69.61 & 50.05 & 19.73 & 17.40 & 12.83 & 71.73\\
\hline
\textsc{Uniter}$_\text{B}$ &  71.29 & 63.95 & 59.48 & 56.31 & 73.33 & 56.80 & 16.53 & 16.93 & 9.74 & 77.46\\
\textsc{Mango}$_\text{B}$ &  72.66 & 66.03 & 61.92 & 58.95 & 74.20 & 58.33 & 15.88 & 16.76 & 9.04 & 78.60\\
\hline
\textsc{Villa}$_\text{B}$ &  72.18 & 65.28 & 60.99 & 57.93 & 73.63 & 58.29 & 15.34 & 17.08 & 9.30 & 79.17\\
\textsc{Mango}$_\text{VB}$ & 72.78 & 65.97 & 61.70 & 58.59 & 74.41 & 58.94 & 15.47 & 16.59 & 9.00 &79.20\\
\hline
\textsc{Uniter}$_\text{L}$ &  74.44 & 67.93 & 63.85 & 60.86 & 72.99 & 57.64 & 15.35 & 17.54 & 9.47 &79.01\\
\textsc{Mango}$_\text{L}$ & \textbf{75.20} & \textbf{69.21} & \textbf{65.38} & \textbf{62.58} & 76.91 & 62.14 & 14.71 & 15.40 & 7.74 &80.86\\
\hline
\textsc{Villa}$_\text{L}$ &  74.93 & 68.65 & 64.61 & 61.61 & 76.18 & 62.00 & \textbf{14.19} & 15.72 & 8.10 &  \textbf{81.38}\\
\textsc{Mango}$_\text{VL}$ & 75.17 & 69.01 & 65.07 & 62.16 & \textbf{77.20} & \textbf{62.60} & 14.60 & \textbf{15.13} & \textbf{7.67} & 81.09\\
\hline
\end{tabu}
\vspace{-8pt}
\caption{\small{Results of consistency evaluations on VQA-Rephrasings and VQA-Introspect. VQA-Reasoning (VQA-Reas.) is a split of VQA-Introspect, containing only the main reasoning questions (M). S stands for sub-questions. $\checkmark$ or $\times$ indicate a correct or wrong prediction. }}
    \label{tab:consistency}
    \vspace{-3mm}
\end{table*}

\noindent\textbf{Robustness against Linguistic Variation} \, As shown in Table~\ref{tab:sota_res} (`Lingual' column), the joint embedding learned by \textsc{Uniter}$_\text{B}$ has shown its advantage of defending model robustness against linguistic variation. We contribute the performance lift from \textsc{Uniter}$_\text{B}$ to excessive variations of textual inputs seen during pre-training. Comparing AT-enhanced methods, \textsc{Mango}$_\text{VB}$ improves over \textsc{Villa}$_\text{B}$, even though \textsc{Villa}$_\text{B}$ has already shown significant improvement over \textsc{Uniter}$_\text{B}$. We attribute the improvement from \textsc{Mango} to not only the adversarial data augmentation during training, but also the random masking introduced from the text modality. More analyses on each component of \textsc{Mango} over VQA-Rephrasings can be found in Table~\ref{tab:ablate}.

\vspace{5pt}
\noindent\textbf{Robustness against Logical Reasoning} \, We compare model performance on 4 benchmarks under the `Reason' column in Table~\ref{tab:sota_res}.  
Different from VQA-LOL Compose, VQA-LOL Supplement dataset consists of questions generated by heuristic rules. Semantically-close questions with different answers are included to make the task more challenging. The close-to-random performance on VQA-LOL Supplement dataset indicates that \textsc{Uniter}$_\text{B}$ severely suffers from these challenging semantically-close questions. 

\textsc{Villa}$_\text{B}$ brings performance lift on all 4 reasoning benchmarks. Not surprisingly, \textsc{Villa}$_\text{B}$ exhibits more robustness than \textsc{Uniter}$_\text{B}$ on semantically-close questions in VQA-LOL Supplement. Our hypothesis is that the adversarial embeddings learned during \textsc{Villa}$_\text{B}$ training can mimic the effect of adding semantically-close questions as training data, and the generated adversarial perturbations are also constrained to be small to preserve the semantic meaning of the clean text embeddings.

\textsc{Mango}$_\text{VB}$ outperforms \textsc{Villa}$_\text{B}$ on all reasoning benchmarks. Similar to VQA-Rephrasings, \textsc{Mango}$_\text{VB}$ has more advantages over VQA-LOL Compose and VQA-LOL Supplement, whose average question length is much longer than VQA v2. By randomly inserting \texttt{[MASK]} tokens, \textsc{Mango}$_\text{B}$ effectively augments training data with questions of similar lengths to the test split.

\vspace{5pt}
\noindent\textbf{Robustness against Visual Content Manipulation} \, \textsc{Uniter}$_\text{B}$ performs on par to SOTA model on IV-VQA, and significantly improves over SOTA on CV-VQA (Table~\ref{tab:sota_res} `Visual' column). This is due to that during pre-training, \textsc{Uniter}$_\text{B}$ has already be trained on diverse images, and the pre-training task of masked region modeling can also prevent \textsc{Uniter}$_\text{B}$ from overfitting to visual biases. \textsc{Villa}$_\text{B}$ improves model robustness against visual content manipulation, and  \textsc{Mango}$_\text{VB}$ performs on par with \textsc{Villa}$_\text{B}$. Our hypothesis is that by injecting adversarial perturbations at pre-training stage, the model is exposed to even more diverse images, hence easier to recover from visual biases.

\vspace{5pt}
\noindent\textbf{Robustness against Answer Distribution Shift} \, On out-of-distribution (OOD) benchmarks, \textsc{Uniter}$_\text{B}$ performs poorly on VQA-CP v2, while improving over SOTA model on GQA-OOD (Table~\ref{tab:sota_res} `Answer' column). As mentioned in Sec.~\ref{sec:main_results}, MUTANT is a very task-specific method, which augments VQA-CP v2 training with excessive rule-based image-question pairs to counter the training split bias. Hence, it is difficult to generalize to other robustness cases. Additional manual effort is required to generalize to other rule-based datasets such as VQA-LOL, GQA, IV-VQA and CV-VQA.
Interestingly, \textsc{Villa}$_\text{B}$ improves over \textsc{Uniter}$_\text{B}$ on GQA-OOD, but not on VQA-CP v2, which may be due to the fact that the generated ``local'' perturbations in \textsc{Villa} cannot cast a strong enough regularization effect on this challenging dataset.
We also observe that \textsc{Mango}$_\text{VB}$ significantly outperforms \textsc{Villa}$_\text{B}$ on both benchmarks, indicating that the generated ``global'' perturbations in \textsc{Mango} are more generalizable to challenging OOD datasets.

\begin{table*}[!t]
\begin{minipage}{.67\textwidth}
    \centering
    \small
\resizebox{.99\textwidth}{!}{
 \begin{tabu}{>{\small}l>{\small}p{0.01\textwidth} >{\small}l c ccccc}
\hline
\rowfont{\small}%
\multirow{2}{*}{Modality} & &  \multirow{2}{*}{Method}  & 
\specialcell{VQA-\\Rep.} & \specialcell{VQA-LOL\\Comp.} & \specialcell{VQA-LOL\\Supp.} & \specialcell{IV-\\VQA} & \specialcell{VQA-CP\\ v2} \\
\cmidrule(lr){4-4} \cmidrule(lr){5-5} \cmidrule(lr){6-6}   \cmidrule(lr){7-7} \cmidrule(lr){8-8} 
\rowfont{\footnotesize}%
& & & Acc. \textcolor{green}{$\uparrow$} & Acc. \textcolor{green}{$\uparrow$} & Acc. \textcolor{green}{$\uparrow$} & \#flips \textcolor{red}{$\downarrow$} & Acc. \textcolor{green}{$\uparrow$}  \\
\hline
 None & 1 & None& 64.56 & 54.54 & 50.00 & 8.47  & 47.29 \\
\hline
\multirow{3}{*}{Image}  & 2 & GN  & 65.17 & 54.46 & 50.68 & 8.45 & 47.29 \\
&3& AN  & 65.42 & 54.59 & 52.54 & 7.52 & 47.38\\

& 4& \textsc{Mango}   & 65.51 & \textbf{56.67} & 55.20 & \underline{7.39}& \underline{47.51}\\
\hline
\multirow{3}{*}{Text} & 5  & GN & 64.73 & 53.66 & 54.59 & 8.46 & 46.59 \\
& 6& AN  & 65.36 & 54.12 & 52.95 & 7.99 & 47.09\\
& 7& \textsc{Mango}   & \underline{65.63} & 55.79 & \textbf{56.54} & 7.53 & 47.45\\
\hline
Both & 8& \textsc{Mango}   & \textbf{65.80} & \underline{56.22} & \underline{56.49} & \textbf{7.32} & \textbf{47.52} \\
\hline
\end{tabu}
}
\vspace{-6pt}
     \caption{\small{Ablation studies on adding noise to different modalities and on different types of noise. \textsc{Uniter}$_\text{B}$ is used as the backbone. GN (AN) stands for Gaussian (Adversarial) Noise.}}
    \label{tab:ablate}
\end{minipage}\hfill
\begin{minipage}{.3\textwidth}
    \centering
    \small
\begin{tabu}{ >{\small}l|ll}
\hline
\rowfont{\small}
Task & LXMERT & Ours\\
\hline
\\[-.7em]
VQA-Rep.& 67.20 & \textbf{68.61}\\
\hline
\\[-.7em]
\specialcelll{VQA-LOL\\Comp.} & 49.34 & \textbf{53.83}\\
\hline
\\[-.7em]
\specialcelll{VQA-LOL\\Supp.} &  47.33 & \textbf{53.54}\\
\hline
\\[-.7em]
GQA  & 59.78 & \textbf{60.06}\\
\hline
\\[-.7em]
GQA-OOD & 53.86 & \textbf{54.94}\\
\hline
\\[-1em]
VQA v2 & 72.31 & \textbf{72.70}\\
\hline
\end{tabu}

\vspace{-6pt}
\caption{\small{Results of \textsc{Mango} with LXMERT as the backbone.}}
    \label{tab:lxmert}
\end{minipage}
\vspace{-3mm}
\end{table*}

\vspace{5pt}
\noindent\textbf{Evaluation on Consistency}\,
In addition to accuracy, many benchmarks consider consistency as an additional measure for evaluating model robustness. Here, we take VQA-Rephrasings and VQA-Introspect as examples to demonstrate that \textsc{Mango} can also help boost consistency in model predictions.  Results are summarized in Table~\ref{tab:consistency}. 

On VQA-Rephrasings, we investigate consistency in model predictions on different variants of semantically equivalent questions.  Consistency is measured by a Consensus Score $CS(k)$.\footnote{ Consensus Score is the ratio of the number of subsets where all the answers are correct and the total number of subsets of size $k$. For every group $Q$ with $n$ rephrasings, all subsets of size $k$ are sampled.  The answer to a question is considered correct if it has a non-zero VQA accuracy.} \textsc{Mango} achieves universal performance lift across all consistency measures, compared to each baseline model. The best results are achieved by \textsc{Mango}$_\text{L}$, surpassing SOTA by \emph{+9.43}, \emph{+12.27}, \emph{+13.62}, \emph{+14.40} on $CS(k), k = 1,2,3,4$, respectively.

On VQA-Introspect, we examine consistency between the main reasoning questions and perceptual sub-questions, measured by 5 metrics.
Similarly, \textsc{Mango} brings universal consistency improvements across all baseline models. The best performance is achieved by \textsc{Mango}$_\text{VL}$, surpassing SOTA by \emph{+12.55}, \emph{+5.54}, \emph{+2.27}, \emph{+5.16}, \emph{+10.10} on M$\checkmark$S$\checkmark$, M$\checkmark$S$\times$, M$\times$S$\checkmark$, M$\times$S$\times$, and S$\checkmark|\text{M}\checkmark$, respectively.

\subsection{Ablation Study}\label{sec:ablate}

\noindent\textbf{Noise Generation and Random Masking}\,
We select one dataset from each robustness type as a representative benchmark for ablation studies: VQA-CP v2, VQA-Rephrasings, VQA-LOL (Compose and Supplement), and IV-VQA. Results are summarized in Table~\ref{tab:ablate}. 
First, we compare with the baseline (Sec.~\ref{sec:baseline}) that simply adds Gaussian noise to either image or text modality.\footnote{In our experiments, we set standard deviation to 0.5, and only perturb 50\% of training data via Gaussian noise within each minibatch.} Different from observations in \cite{rusak2020simple}, comparing L2/L5 with L1 indicates that adding simple Gaussian noise to multimodal embeddings is not always helpful. Especially, adding Gaussian noise on text modality brings unstable performance. 

Second, we experiment with adding adversarial noise alone, without random masking. Results on L3/L6 show that universal performance improvements over Gaussian noise (L2/L5). Intuitively, adversarial noise is \emph{harder} than Gaussian noise, as the adversarial noise generator learns to fool the backbone network. Model training with such augmented hard examples helps to boost model robustness. 


Third, we show that by using random masking (L4/L7), which encourages more diverse adversarial embeddings, \textsc{Mango} is better than using adversarial noise alone (L2/L5). 
Randomly inserting \texttt{[MASK]} tokens (L7) also shifts the distribution of question lengths that the model is exposed to during training. Hence, we observe more gains on benchmarks with severe mismatches in question length between training and test sets.
For example, in VQA-LOL, the questions in the test set are significantly longer than questions in the training set on average.

Lastly, we observe that adding adversarial noise on one modality is already gaining significant improvement (L4/L7). Empirically, adding adversarial noise on both modalities (L8) only performs slightly better or on par with \textsc{Mango} on text or image modality alone. More ablation results on model architecture are included in Appendix.
\begin{table}[t!]
\centering
\small
\begin{tabular}{ l  c c c c}
  \hline			
   Model &  NLVR$^2$  & RefCOCO & RefCOCOg & VE \\
  \hline
  \textsc{Uniter}$_{\text{B}}$ & 77.52 & 80.55 & 74.41& 78.44 \\
  \textsc{Mango}$_{\text{B}}$ & \textbf{78.36} &  \textbf{80.95} &  \textbf{75.37} & \textbf{78.87}\\
  \hline
\end{tabular}
\vspace{-2mm}
\caption{\small{Results on other V+L tasks, we report the average of performance across different splits of each task for simplicity.
}}
\label{tab:other}
\vspace{-4mm}
\end{table}
\vspace{5pt}
\noindent\textbf{Results on LXMERT}\,
To demonstrate the versatility of \textsc{Mango}, we also apply \textsc{Mango} to a two-stream backbone, LXMERT, for generalizability test. We compare LXMERT baseline with its enhanced version with \textsc{Mango} (``ours'' in Table~\ref{tab:lxmert}). Evaluation is conducted over VQA-Rephrasings, VQA-LOL Compose, VQA-LOL Supplement, GQA and GQA-OOD datasets, covering 3 types of robustness. IV-VQA, CV-VQA and VQA-CP v2 are excluded in this study as the performance on these benchmarks is based on examples in VQA v2 val split, which is used to supervise LXMERT pre-training.   We also report results on standard VQA v2 benchmark. 

Note that LXMERT is pre-trained with VQA v2 and GQA data. Therefore, it achieves superior performance on VQA-Rephrasings, which includes questions from VQA dataset. On benchmarks whose data are unseen during pre-training, LXMERT exhibits similar robustness to \textsc{Uniter}. LXMERT suffers severely on semantically-close questions in VQA-LOL Supplement and logical reasoning questions in VQA-LOL Compose. A possible reason is the over-exposure to VQA questions during both pre-training and finetuning.  Despite the limitations mentioned above, when applying \textsc{Mango} to LXMERT, we still observe universal performance lift across all benchmarks considered.

\vspace{5pt}
\noindent\textbf{Results on other V+L tasks}\, As aforementioned in Sec.~\ref{sec:benchmark}, we take robust VQA as test bed (9 datasets with 10 different model settings) due to the lack of available datasets to evaluate model robustness for other V+L tasks. \textsc{Mango} is task-agnostic, thereby can also be applied to other standard V+L tasks. We compare \textsc{Mango}$_\text{B}$ against \textsc{Uniter}$_\text{B}$ on popular V+L tasks, including NLVR$^2$~\cite{suhr2018corpus}, RefCOCO~\cite{yu2016modeling}, RefCOCOg~\cite{yu2016modeling} and Visual Entailment (VE)~\cite{xie2019visual} in Table~\ref{tab:other}.  \textsc{Mango} surpasses baseline results across all 4 tasks. 

\begin{figure}
    \centering
    \includegraphics[width=0.45\textwidth]{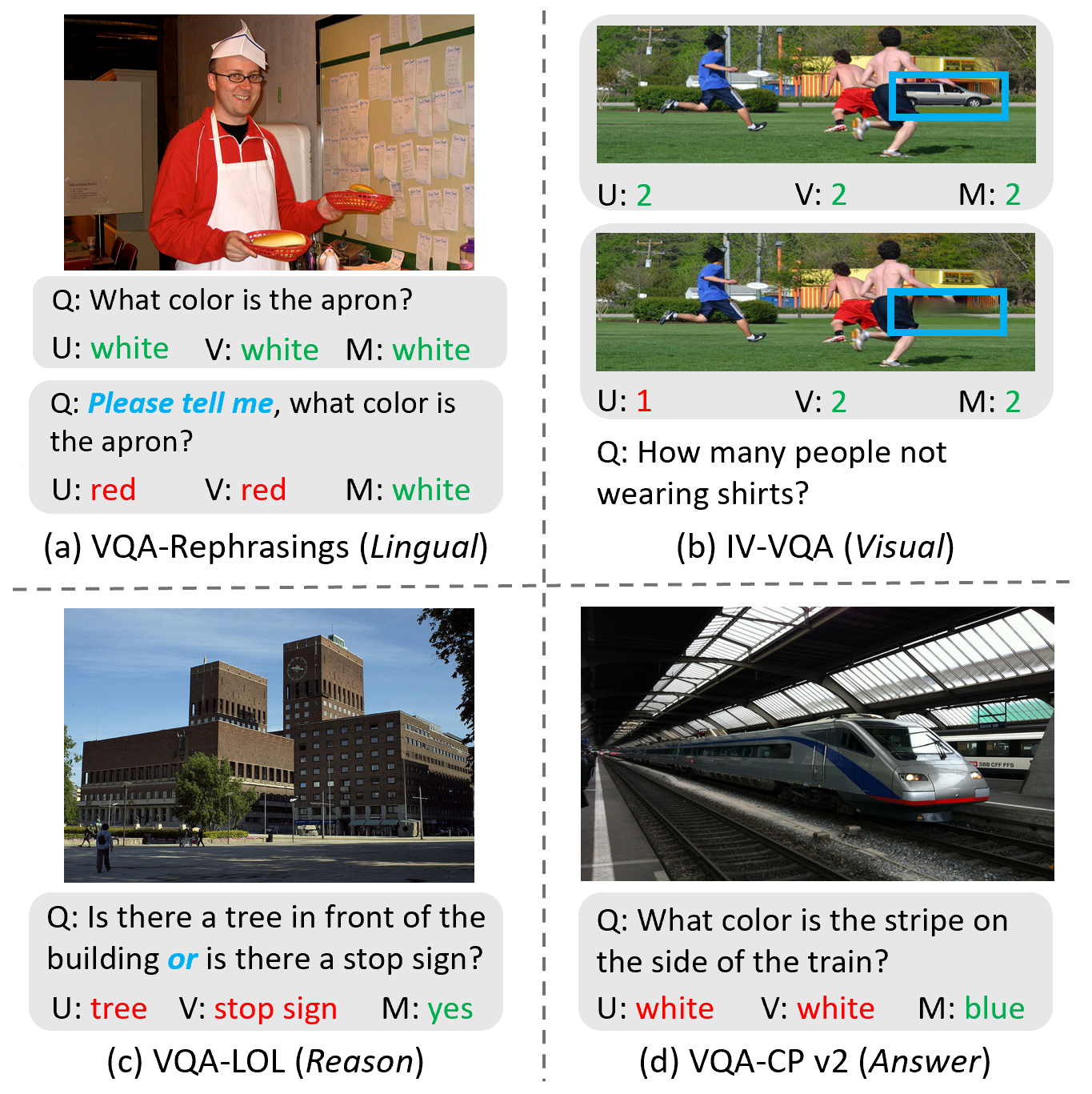}
    \vspace{-3mm}
    \caption{\small Visualization of model predictions, comparing \textsc{Mango}(M) against  \textsc{Uniter}(U) and \textsc{Villa}(V). Correct answers are highlighted in \textcolor{ForestGreen}{green} and wrong ones are in \textcolor{red}{red}. }
    \label{fig:vis}
    \vspace{-4mm}
\end{figure}

\vspace{5pt}
\noindent\textbf{Qualitative Analysis}\, Figure~\ref{fig:vis} visualizes prediction examples from \textsc{Uniter}, \textsc{Villa} and \textsc{Mango} on 4 benchmarks (each for one robustness type). These visualizations illustrate \textsc{Mango}'s consistently accurate performance when facing challenges of: (a) uninformative leading phrase added to the question; (b) removal of irrelevant object in the image; (c) over-length logical combination of questions; and (d) imbalanced answer distribution (`white' appears 3 times as many as `blue' in training set). More visualization examples are included in Appendix.

\vspace{-0.5mm}
\section{Conclusion}
\vspace{-0.5mm}
We provide the first known systematic study on the robustness of pre-trained V+L models. By examining existing models on a wide range of robustness benchmarks, we obtain a better understanding of how V+L pre-training handles various types of robust tests. We further propose \textsc{Mango}, a simple yet efficient adversarial training method to enhance model robustness, which advances the state of the art on 7 out of 9 robustness benchmarks by a large margin. We hope this set of results can be used as baseline for future research. A natural follow-up of this work is to investigate the \emph{adversarial} robustness of pre-trained V+L models.


\clearpage
{\small
\bibliographystyle{ieee_fullname}
\bibliography{egbib}
}
\clearpage
\appendix
\section{Detailed Related Work}\label{sec:related_work}
\paragraph{Multimodal Pre-training}
Early approaches to vision-and-language pre-training~\cite{lu2019vilbert,tan2019lxmert} adopt a two-stream architecture. 
Later on, single-stream architecture gains popularity~\cite{su2019vl,li2019visualbert,alberti2019fusion,li2019unicoder,chen2019uniter,lin2020interbert}. 
Multi-task learning~\cite{lu201912}, adversarial training~\cite{gan2020large}, and contrastive learning~\cite{shi2020contrastive} have proved useful for improving model performance. 
Probing analysis~\cite{cao2020behind} shows that these pre-trained models can learn essential knowledge about visual co-reference and visual relations. Recent work extends this pre-training strategy to diverse tasks such as image captioning~\cite{zhou2019unified,hu2020vivo}, visual dialog~\cite{murahari2019large,wang2020vd}, visual-language navigation~\cite{hao2020towards,majumdar2020improving}, and video-text pre-training~\cite{sun2019videobert,sun2019learning,li2020hero,zhu2020actbert}. 
Instead of using the conventional bottom-up-attention features~\cite{anderson2018bottom}, Pixel-BERT~\cite{huang2020pixel} proposes end-to-end learning from image pixels to textual tokens. External knowledge, such as image tags~\cite{li2020oscar} and scene graphs~\cite{yu2020ernie}, as well as weakly-supervised pre-training~\cite{li2020weakly}, are also investigated for further enhancement. 

Distinct from these efforts on improving performance over standard benchmarks,\footnote{Examples of standard benchmarks include VQA~\cite{antol2015vqa}, VCR~\cite{zellers2019recognition}, NLVR$^2$~\cite{suhr2018corpus}, Image-Text Retrieval~\cite{lee2018stacked}, and Referring Expressions~\cite{yu2016modeling}.}
we focus on a different direction, evaluating and enhancing the \emph{robustness} of pre-trained models. This helps us better understand how well multimodal pre-training truly advances this field, and guides us to design more robust models.



\vspace{5pt}
\noindent\textbf{Adversarial Training}\,
As one of the most effective strategies of defending against adversarial attacks~\cite{szegedy2013intriguing,goodfellow2014explaining,athalye2018obfuscated}, adversarial training (AT) has been widely studied for enhancing adversarial robustness of neural networks~\cite{tramer2017ensemble,madry2017towards,xie2019feature,shafahi2019adversarial,pang2020boosting,wong2020fast, zhang2019theoretically, xie2020adversarial}, using adversarial examples as effective data augmentation.
Recent studies show that, by injecting adversarial perturbations into feature space, AT can further improve model generalization on language understanding~\cite{zhu2019freelb,jiang2019smart,liu2020adversarial}, visual question answering~\cite{gan2020large,tang2020semantic}, and graph neural networks~\cite{kong2020flag}.

In our work, we investigate the use of an adversarial noise generator for robustness enhancement, inspired by \cite{rusak2020simple}, which proposes a similar noise generator to make neural networks robust
against diverse image corruptions. However, there are two key distinctions. First, we focus on transformer models designed for V+L tasks, whereas \cite{rusak2020simple} focuses on convolutional networks for image classification. Second, we propose to generate adversarial noise over the embeddings of images and words, while \cite{rusak2020simple} adds adversarial noise directly on image pixels.

\section{More Results}
We include more detailed results on  IV-VQA~\cite{agarwal2020causal-vqa}, CV-VQA~\cite{agarwal2020causal-vqa}, GQA-OOD~\cite{kervadec2020gqa-ood} and VQA v2~\cite{goyal2017making}, and also ablation experiments on model architecture. 

\vspace{5pt}
\noindent\textbf{On IV-VQA and CV-VQA}, we decouple the inconsistency in model predictions on edited images (measured by \#flips) into 3 categories: ($i$) p2n: answer predicted on
the edited image was wrong, but the prediction on the corresponding real image was correct; ($ii$) n2p: model makes a correct prediction on the edited image, while predicting a wrong answer on real image; ($iii$) n2n: different answers were predicted on edited and real images and both are wrong.  These metrics  may expose that there is brittleness even when the model makes correct predictions, indicating that models often exploit spurious correlations while making predictions. We follow~\cite{agarwal2020causal-vqa} to report accuracy on VQA v2 val split to serve as reference for IV-VQA, and performance on counting questions in VQA v2 val split for CV-VQA.

Similar conclusions are drawn from the results presented in Table~\ref{tab:inv_vqa}. First, \textsc{Mango} brings consistent performance improvements across all metrics on both benchmarks, compared to \textsc{Uniter}. Second, \textsc{Mango} significantly improves over SOTA. We also observe significant improvements from \textsc{Mango} over \textsc{Villa} on CV-VQA. These results suggest that for \emph{challenging} questions such as counting problems in CV-VQA, \textsc{Mango} is more robust than \textsc{Villa}.

\begin{table*}[ht]
    \centering
\resizebox{\textwidth}{!}{
\begin{tabu}{>{\small}lcccccc||cccccc}
\hline
\rowfont{\small}%
\multirow{2}{*}{ \normalsize{Model} } & VQA & \multicolumn{5}{c}{IV-VQA} & VQA Num. & \multicolumn{5}{c}{CV-VQA} \\
\cmidrule(lr){2-2} \cmidrule(lr){3-7} \cmidrule(lr){8-8} \cmidrule(lr){9-13}
\rowfont{\small}%
& Acc.\textcolor{green}{$\uparrow$} & Acc.\textcolor{green}{$\uparrow$} & \# of flips \textcolor{red}{$\downarrow$} & p2n \textcolor{red}{$\downarrow$} & n2p \textcolor{red}{$\downarrow$} & n2n \textcolor{red}{$\downarrow$} & Acc.\textcolor{green}{$\uparrow$} & Acc.\textcolor{green}{$\uparrow$} & \# of flips \textcolor{red}{$\downarrow$} & p2n \textcolor{red}{$\downarrow$} & n2p \textcolor{red}{$\downarrow$} & n2n \textcolor{red}{$\downarrow$}\\
\hline
SOTA~\cite{kazemi2017show} & 70.26 & - & 7.85 & 3.47 & 2.79 & 1.58 & 49.9 & - & 78.44 & 31.66 & 25.38 & 21.40\\
\hline
\textsc{Uniter}$_\text{B}$  & 70.34 & 83.35 & 8.47 & 3.89 & 2.60 & 1.97 & 53.82 & 63.22 & 40.67 & 23.21 & 10.72 & 6.74\\
\textsc{Mango}$_\text{B}$  & 71.17 & 82.69 & 7.32 & 3.55 & 2.27 & 1.49 & 54.86 & 64.21 & 38.11 & 22.22 & 9.97 & 5.92 \\
\hline
\textsc{Villa}$_\text{B}$ & 71.27 & 82.87 & 7.07 & 3.48 & 2.16 & 1.44 & 55.02 & 65.06 & 38.28 & 22.17 & 10.60 & 5.51\\
\textsc{Mango}$_\text{VB}$  & 71.47 & 82.84 & 7.43 & 3.57 & 2.34 & 1.52 & 55.27 & 65.66 & 38.25 & 22.10 & 9.78 & 6.38\\
\hline
\textsc{Uniter}$_\text{L}$ & 72.60 & \textbf{85.86} & 8.20 & 3.96 & 2.37 & 1.88 & 56.61 & 67.13 & 36.66 & \underline{22.05} & 9.80 & 4.81\\
\textsc{Mango}$_\text{L}$  & \underline{73.06} & 84.05 &  \textbf{6.69} &  \textbf{3.34} & 2.00 & \underline{1.34} & \underline{57.44} & \underline{67.30} &  \textbf{35.52} &  \textbf{21.59} &  \textbf{8.55} & 5.39\\
\hline
\textsc{Villa}$_\text{L}$ &  \textbf{73.20} & \underline{84.79} & \underline{6.70} & 3.45 & \underline{1.95} &  \textbf{1.29} & 57.43 & 65.54 & 37.55 & 24.05 & 8.94 &  \textbf{4.56}\\
\textsc{Mango}$_\text{VL}$  & 72.96 &   84.70& 6.73& \underline{3.42} &  \textbf{1.89} & 1.42&  \textbf{57.53} &  \textbf{67.86} & \underline{35.64} & 22.07 & \underline{8.79} & \underline{4.78}\\
\hline
\end{tabu}
}
    \caption{\small{Detailed Results on IV-VQA and CV-VQA. \textcolor{green}{$\uparrow$} (\textcolor{red}{$\downarrow$}) indicate the higher (lower) the better.}}
    \label{tab:inv_vqa}
\end{table*}

\begin{table}[t]
    \centering
    \small
\begin{tabu}{>{\small}lcccc}
\hline
\normalsize{Model} & All \textcolor{green}{$\uparrow$} & Tail \textcolor{green}{$\uparrow$} & Head \textcolor{green}{$\uparrow$} & $\Delta$ \textcolor{red}{$\downarrow$}\\
\hline
SOTA\cite{chen2021meta}(best All) & 52.70 & 48.00 & 55.50 & 15.60 \\
SOTA\cite{Kim2018bilinear} (best $\Delta$) & 50.20 & 47.20 & 51.90 & \textbf{9.90} \\
\hline
\textsc{Uniter}$_\text{B}$ & 53.43 & 48.45 & 56.49 & 16.59 \\
\textsc{Mango}$_\text{B}$  & 54.47 & 50.24 & 57.07 & \underline{13.59} \\
\hline
\textsc{Villa}$_\text{B}$ & 54.11 & 49.86 & 56.72 & 13.76 \\
\textsc{Mango}$_\text{VB}$  & 55.79 & \underline{50.89} & 58.74 & 15.43 \\
\hline
\textsc{Uniter}$_\text{L}$ & 53.65 & 48.82 & 56.61 & 15.96 \\
\textsc{Mango}$_\text{L}$ & \textbf{56.40} & \textbf{51.27} & \textbf{59.55} & 16.15 \\
\hline

\textsc{Villa}$_\text{L}$ & 55.26 & 50.80 & 58.05 & 14.27 \\
\textsc{Mango}$_\text{VL}$  & \underline{56.08} & \textbf{51.27} & \underline{59.03} & 15.14 \\
\hline
\end{tabu}
    \caption{\small{Detailed Results on GQA-OOD. \textcolor{green}{$\uparrow$} (\textcolor{red}{$\downarrow$}) indicate the higher (lower) the better.}}
    \label{tab:gqa_ood}
\end{table}

\vspace{5pt}
\noindent\textbf{On GQA-OOD}, except for the accuracy over all GQA-OOD samples (`All' in Table~\ref{tab:gqa_ood}), three additional metrics are considered: ($i$)  the accuracy on OOD samples, which are the samples of the tail of the answer class distribution (`Tail'); ($ii$) the accuracy on the head of distribution (`Head'); and ($iii$) $\Delta(\text{head, tail}) = (\text{head - tail})/\text{tail}$ to illustrate how much the error prediction is imbalanced between frequent and rare answers (`$\Delta$'). More details on the statistics of head and tail examples can be found in~\cite{kervadec2020gqa-ood}. \textsc{Mango} achieves universal performance lift across all accuracy measures, compared to each baseline model. However, better accuracy does not indicate better-balanced predictions between tail and head splits. We observe that there are more performance improvements on head split than tail split. When compared to SOTA, \textsc{Mango}$_\text{B}$ surpasses MMN~\cite{chen2021meta} (SOTA with the best All) across all metrics. BAN~\cite{Kim2018bilinear} is the SOTA method with the best $\Delta$; however, it suffers on all accuracy measures.

\begin{table}[t!]
\centering
\small
\begin{tabular}{ l  c c c c}
  \hline			
   Method &  All  & Y/N & Num & Other \\
  \hline
  \textsc{Uniter}$_{\text{B}}$ & 72.70 & 88.97 & 55.67 & 62.81\\
  \textsc{Mango}$_{\text{B}}$ & \textbf{73.24} & \textbf{89.27} & \textbf{56.48} & \textbf{63.34}\\
  \hline
\end{tabular}
\caption{\small{Detailed results of \textsc{Uniter}$_{\text{B}}$ and \textsc{Mango}$_{\text{B}}$ on VQA v2.}}
\label{tab:vqa_q_type}
\end{table}
\vspace{5pt}
\noindent\textbf{On VQA v2}, we use \textsc{Mango}$_{\text{B}}$ and \textsc{Uniter}$_{\text{B}}$ as examples to show that our method can provide universal performance lift for each question type. This is also consistent with our observations on various robust vqa benchmarks, as they focus on different question types by design. For examples, IV-VQA speficially desgined for counting questions, VQA-LOL only includes yes/no questions.

\begin{table}[t!]
\centering
\small
\begin{tabular}{lcc}
  \hline			
   Method &  VQA-Rep. & VQA-LOL \\
  \hline
  \textsc{Mango}$_{\text{B}}$ & \textbf{65.80} & \textbf{56.61} \\
  $- L_{kl}$ & 65.01 & 54.55 \\
  $-$ retrain NG & 65.48 & 54.57 \\
  w/ 1-layer linear NG & 65.54 & 53.14 \\
  \hline
  \textsc{Villa}$_{\text{B}}$ + Masking & 65.46 &  55.96\\
  \textsc{Mango}$_{\text{VB}}$ & \textbf{65.91} &  \textbf{56.55}\\
  \hline
\end{tabular}
\caption{\small{Additional ablation results.
}}
\label{tab:more_ablation}
\end{table}
\vspace{5pt}
\noindent\textbf{Additional Ablations} We conduct additional ablation studies to validate several model design choices, including KL-divergence Loss, retraining noise generator every $T$ steps (retrain NG), the architecture of NG (multiple linear layers with nonlinear activation) and the effectiveness of masking on \textsc{Villa}. Results are reported in Tab.~\ref{tab:more_ablation}. 

A few key observations are summarized here: ($i$) KL divergence loss contributes to performance improvements in \textsc{Mango}. ($ii$) Without resetting generator parameters and retraining generator periodically results in worse performance. As explained in L409-415 of the main text, the lightweight generator may be trapped in a local optima. ($iii$) Replacing our noise generator with a single linear layer also hurts the performance.  Note that applying linear layers to a Gaussian noise only changes its mean and variance, still results in a Gaussian noise. ($iv$) \textsc{Villa}$_{\text{B}}$ + Masking  renders weaker performance than \textsc{Mango}$_{\text{VB}}$. This observation is consistent with comparison of \textsc{Villa}$_{\text{B}}$ in Tab. 2 and ``AN" in Tab. 4, which can be considered as ``\textsc{Mango}$_{\text{B}}$ - Masking''. 

\begin{table*}[t!]
\centering
\small
\begin{tabular}{l c c c c c c c c }
\toprule
Task & Model & Training Steps & $p_{\text{mask}}^{\text{img}}$ & $p_{\text{mask}}^{\text{txt}}$ &\specialcell{Adv. Noise \\Lr.} & \specialcell{kl-div loss\\weight $\beta$} & \specialcell{Adv. Noise\\Retrain steps}  & \specialcell{Adv. Noise\\Retrain Lr.} \\ 
\midrule
\multirow{2}{*}{VQA-Rephrasings} & \textsc{Mango}$_\text{B}$ &  4000 & 0.15 & 0.15 & 1e-5 & 1.0 & 400 &   1e-4 \\
& \textsc{Mango}$_\text{L}$ &3000 & 0.15 & 0.30 & 5e-6 & 1.0 & 400 &   5e-5 \\
\midrule
\multirow{2}{*}{VQA-LOL} & \textsc{Mango}$_\text{B}$ & 4000 & 0.15 & 0.45 & 1e-5 & 1.0 & 400 &   1e-4  \\
& \textsc{Mango}$_\text{L}$ & 4000 & 0 & 0.45 & 5e-6 & 1.0 & 400 &   5e-5\\
\midrule
\multirow{2}{*}{VQA-Introspect} & \textsc{Mango}$_\text{B}$ &2000 & 0 & 0.15 & 1e-5 & 1.0 & 400 &   1e-4   \\
& \textsc{Mango}$_\text{L}$ & 3000 & 0 & 0.45 & 5e-6 & 1.0 & 400 &   5e-5 \\
\midrule
\multirow{2}{*}{GQA} & \textsc{Mango}$_\text{B}$ & 4000 & 0.15 & 0.15 & 1e-5 & 1.0 & 800 &   1e-4  \\
& \textsc{Mango}$_\text{L}$   & 4000 & 0.15 & 0.15 & 1e-5 & 1.0 & 800 &   1e-4\\
\midrule
\multirow{2}{*}{VQA CP v2} & \textsc{Mango}$_\text{B}$ & 3000 & 0.15 & 0.6 & 1e-5 & 0 & 400 &   1e-4 \\
& \textsc{Mango}$_\text{L}$ &3000 & 0.15 & 0.6 & 1e-6 & 0 & 400 &   1e-5  \\
\midrule
\multirow{2}{*}{GQA-OOD} & \textsc{Mango}$_\text{B}$ & 4000 & 0.15 & 0.15 & 1e-5 & 1.0 & 800 &   1e-4 \\
& \textsc{Mango}$_\text{L}$  & 2000 & 0.15 & 0.15 & 1e-5 & 1.0 & 800 &   1e-4 \\
\midrule
\multirow{2}{*}{\specialcell{IV-VQA\\\&CV-VQA}} & \textsc{Mango}$_\text{B}$ & 4000 & 0.15 & 0.45 & 1e-5 & 1.0 & 400 &   1e-4 \\
& \textsc{Mango}$_\text{L}$ &3000 & 0.15 & 0.15 & 5e-6 & 1.0 & 400 &   5e-5 \\
\midrule
\multirow{2}{*}{VQA v2} & \textsc{Mango}$_\text{B}$ & 6000  & 0.15 & 0.45 & 1e-5 & 1.0 & 400 &   1e-4    \\
& \textsc{Mango}$_\text{L}$ & 5000  & 0.15 & 0.45 & 1e-5 & 1.0 & 400 &   1e-4 \\
\bottomrule
\end{tabular}
\caption{\small Hyper-parameter values used in our experiments. We use batch size of 5120 (3072) and gradient accumulation steps of 5 (8) for base (large) model experiments. }
\label{table:hyper_param}
\end{table*}

\section{Implementation Details}
Our models are implemented based on PyTorch.\footnote{https://pytorch.org/} To speed up training, we use Nvidia Apex\footnote{https://github.com/NVIDIA/apex} for mixed precision training. Gradient accumulation~\cite{ott2018scaling} is applied to reduce multi-GPU communication overheads. All experiments are run on Nvidia V100 GPUs (32GB VRAM; NVLink connection). We use AadmW~\cite{loshchilov2017decoupled} with $\beta_1{=}0.9$, $\beta_2{=}0.98$ and an L2 weight decay of 0.01 to optimize model training. Throughout the training, the learning rate is scheduled to warmup over the first 10\% training steps followed by linear decay to 0. The peak learning rate is set to be 8e-5 and 5e-5 for base and large models, respectively. Additional hyper-parameters used to train our adversarial noise generators are listed in Table~\ref{table:hyper_param}. Empirically, we found that model training is sensitive to adversarial noise retrain steps, $p_{\text{mask}}^{\text{img}}$ and $p_{\text{mask}}^{\text{txt}}$.

\section{Downstream Benchmarks}
In addition to dataset statistics summarized in Table 1 in main text,  we provide an overview of each robustness benchmark as follows.

\vspace{5pt}
\noindent\textbf{VQA-Rephrasings}~\cite{shah2019vqa-rephrase}\,is based on VQA v2~\cite{goyal2017making}. It contains 3 human-provided rephrasings for 40K questions on 40K images from VQA v2 val split. In addition to accuracy, consistency in model predictions to different semantically-equivalent questions is also used to measure the robustness of VQA models against linguistic variations. We follow~\cite{shah2019vqa-rephrase} to evaluate models trained with VQA v2 train split.

\vspace{5pt}
\noindent\textbf{VQA-LOL}~\cite{gokhale2020vqa-lol}\, is introduced to examine the logical reasoning ability of a VQA model through questions containing logical compositions and linguistic transformations (negation, disjunction,
conjunction, and antonyms). It consists of two datasets: VQA-LOL \emph{Compose} (logical combinations of multiple closed binary questions about the same image in VQA v2) and VQA-LOL \emph{Supplement} (logical combinations of additional questions based on external object and caption annotations about the images from COCO~\cite{chen2015microsoft}). Both datasets share the same train/val images as VQA v2. In total, 757K/42.5K/291K and 1.61M/91.8K/669K image-question pairs are generated for train/val/test splits of VQA-LOL Compose and VQA-LOL Supplement, respectively. In our experiments, we follow~\cite{gokhale2020vqa-lol} to evaluate models trained with VQA v2 train split on test split of both datasets.

\vspace{5pt}
\noindent\textbf{VQA-Introspect}~\cite{selvaraju2020vqa-introspect}\ is created to investigate the consistency in model predictions of a VQA model between reasoning questions and their associated low-level perception questions. It first introduces a new Reasoning split of the VQA v2 dataset and collects 238K new perception questions. These questions correspond to the set of perceptual tasks needed to effectively answer complex reasoning questions in the Reasoning split. In total, VQA-Introspect contains 167K sub-questions for 56K reasoning questions in VQA v2 train, and 72K sub-questions for 22K reasoning questions in VQA v2 val. In our experiments, we follow~\cite{selvaraju2020vqa-introspect} to evaluate models trained with VQA v1~\cite{antol2015vqa} train split on VQA-Introspect val split.

\vspace{5pt}
\noindent\textbf{GQA}~\cite{hudson2019gqa}\, contains 22M automatically generated questions based on ground-truth image scene graphs. The questions are constructed via a set of heuristic rules, which are designed to evaluate a VQA model in terms of different types of reasoning skills (\emph{e.g.}, spatial understanding and multi-step inference). We follow~\cite{hudson2019nsm} to use the balanced version of GQA, which has been designed to reduce biases in answer distribution. In the balanced version, 1.7M questions are split into 70\%/10\%/10\% for training, validation and test sets, respectively. In our experiments, models are trained on GQA train split and we report performance on test-dev split. 

\vspace{5pt}
\noindent\textbf{IV-VQA \& CV-VQA}~\cite{agarwal2020causal-vqa}\, are two synthetic datasets, created by removing objects in the real VQA images. In IV-VQA, irrelevant objects are erased and model predictions before and after image manipulations are expected to be invariant. In CV-VQA, which focuses on counting questions, one relevant object is removed from the given image and model predictions on the quantity of such object are expected to be subtracted by 1.  Objects of choice are based on heuristic rules and removed via inpainter-GAN~\cite{shetty2018adversarial}. In total, 376K and 13K image-question pairs are generated for IV-VQA and CV-VQA, respectively. The detailed splits can be found in Section 3 of main text.  In our experiments, we follow~\cite{agarwal2020causal-vqa} to evaluate models trained with VQA v2 train split on IV-VQA/CV-VQA val split.

\begin{figure*}
    \centering
    \includegraphics[width=\linewidth]{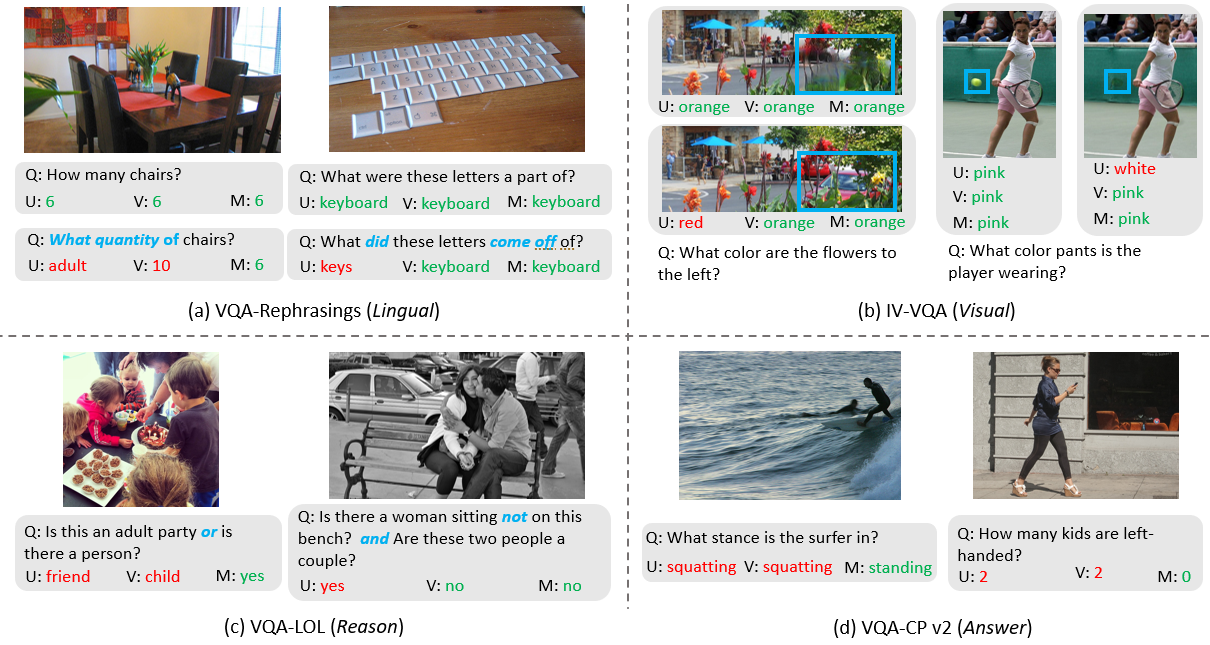}
    \caption{\small More visualization of model predictions, comparing \textsc{Mango} (M) against  \textsc{Uniter} (U) and \textsc{Villa} (V). Correct answers are highlighted in \textcolor{ForestGreen}{green} and wrong ones are in \textcolor{red}{red}. }
    \label{fig:more_vis}
\end{figure*}
\vspace{5pt}

\vspace{5pt}
\noindent\textbf{VQA-CP v2}~\cite{agrawal2018vqa-cp}\, is an out-of-distribution (OOD) reorganization of VQA v2. It was
created to examine the robustness of a VQA model in a setting where language priors cannot be relied upon for a correct prediction.  The questions in VQA v2 are first assigned to one of 65 question types according to their prefix (first few words). For every question type, the prior distribution of answers is shuffled to be different in train and test splits of VQA-CP v2. Our models are trained on VQA-CP v2 train split and evaluated on test split, following~\cite{agrawal2018vqa-cp}.

\noindent\textbf{GQA-OOD}~\cite{kervadec2020gqa-ood}\, is also an OOD benchmark, created by re-organization of the GQA dataset. By utilizing fine-grained question generation templates in GQA, GQA-OOD divides questions into 37K local groups, and shifts answer distribution by selecting a subset of answer classes for each question group, according to their frequencies. Unlike VQA-CP v2, GQA-OOD  features distribution shifts for both validation and test, allowing to validate models under OOD conditions. In our experiments, we follow~\cite{kervadec2020gqa-ood} to evaluate models trained with GQA train split on GQA-OOD test-dev split.

\section{More Visualizations}
We provide additional visualization of model predictions in Figure~\ref{fig:more_vis}. \textsc{Mango} consistently provides accurate predictions for each robustness type. 

\end{document}